\documentclass{article}
\usepackage{multirow} 
\usepackage[table]{xcolor} 
\usepackage{xcolor}
\newcommand{\gain}[1]{\textcolor{green!60!black}{\small #1}}
\usepackage{microtype}
\usepackage{graphicx}
\usepackage{subcaption}
\usepackage{booktabs} 
\usepackage{tikz}
\usetikzlibrary{shapes.geometric}
\usepackage{xspace}
\newcommand{\eg}{\textit{e.g.}\@\xspace}
\newcommand{\ie}{\textit{i.e.}\@\xspace}
\newcommand{\capcirc}{\tikz[baseline=-0.6ex]\draw[fill=white, draw=black, line width=0.35pt] (0,0) circle (0.9ex);}
\newcommand{\capstar}{\tikz[baseline=-0.6ex]\node[star, star points=5, star point ratio=2.25, fill=white, draw=black, line width=0.35pt, inner sep=0.35ex] {};}

\usepackage{hyperref}

\newcommand{\mcshade}{\cellcolor{black!8}}

\usepackage{algorithm}
\usepackage{algorithmic} 

\usepackage[preprint]{icml2026}

\usepackage{amsmath}
\usepackage{amssymb}
\usepackage{mathtools}
\usepackage{amsthm}

\usepackage[capitalize,noabbrev]{cleveref}

\theoremstyle{plain}

\theoremstyle{definition}

\theoremstyle{remark}

\usepackage[textsize=tiny]{todonotes}

\icmltitlerunning{MC-GRPO: Median-Centered Group Relative Policy Optimization  for Small-Rollout Reinforcement Learning}

\begin{document}

\twocolumn[
  \icmltitle{MC-GRPO: Median-Centered Group Relative Policy Optimization \\ for Small-Rollout Reinforcement Learning }



  \icmlsetsymbol{equal}{*}

  \begin{icmlauthorlist}
    \icmlauthor{Youngeun Kim \textsuperscript{*}}{a}
  \end{icmlauthorlist}

  \icmlaffiliation{a}{Amazon}

  \icmlcorrespondingauthor{Youngeun Kim}{youngeun.ryan.kim@gmail.com}

  \icmlkeywords{Machine Learning, ICML}

  \vskip 0.3in
]



\printAffiliationsAndNotice{\textsuperscript{*} Most of this work was completed prior to joining Amazon.}

\begin{abstract}
  Group-relative policy optimization methods train language models by generating multiple rollouts per prompt and normalizing rewards with a shared mean reward baseline. In resource-constrained settings where the rollout budget is small, accuracy often degrades. We find that noise in the shared baseline induces advantage sign flips, where some rollouts receive an incorrect advantage sign, and the update direction is reversed. To address this, we propose Median-Centered Group Relative Policy Optimization (MC-GRPO), a simple and effective solution for small-rollout training. 
  Our main idea is to replace the mean baseline with a median baseline: the median is far less sensitive to outlier rewards than the mean, mitigating the sign flips under small rollout size ($G$). 
  We generate one additional rollout for median reference ($G+1$), and compute advantages by using the group median. With an odd-sized group, exactly one completion is the median and receives zero advantage, we exclude this pivot rollout from backpropagation so the number of gradient-contributing samples per prompt remains $G$, preserving the core update cost of standard  $G$-rollout training. Across various GRPO-family methods and a wide range of models and scales, this median-centered training consistently improves stability and final accuracy in the low-rollout regime, reducing the gap between $G{=}2$ and $G{=}8$ to within $1\%$. Code is available at \href{https://github.com/lotusroot-kim/MC-GRPO}{\texttt{lotusroot-kim/MC-GRPO}}.
\end{abstract}

\section{Introduction}
Online reinforcement learning (RL) for large language models (LLMs) has recently emerged as an effective paradigm for directly optimizing task objectives from rule-based rewards, learned reward models, or preference signals~\cite{schulman2017ppo,christiano2017preferences,ouyang2022instructgpt,bai2022constitutional,guo2025deepseek,team2025kimi}.
A common practice in this setting is to sample multiple responses (\ie, rollouts) per prompt and learn from their {relative} quality.
Group-relative methods such as GRPO~\cite{shao2024deepseekmath} instantiate this idea by computing advantages within each prompt group using a shared baseline.
Given rewards $\{r_i\}_{i=1}^{G}$ for $G$ rollouts, GRPO subtracts the within-prompt mean reward to produce a centered, relative learning signal that has been shown to improve training stability and downstream accuracy.

\begin{figure}[t]
  \centering
  \includegraphics[width=0.85\linewidth]{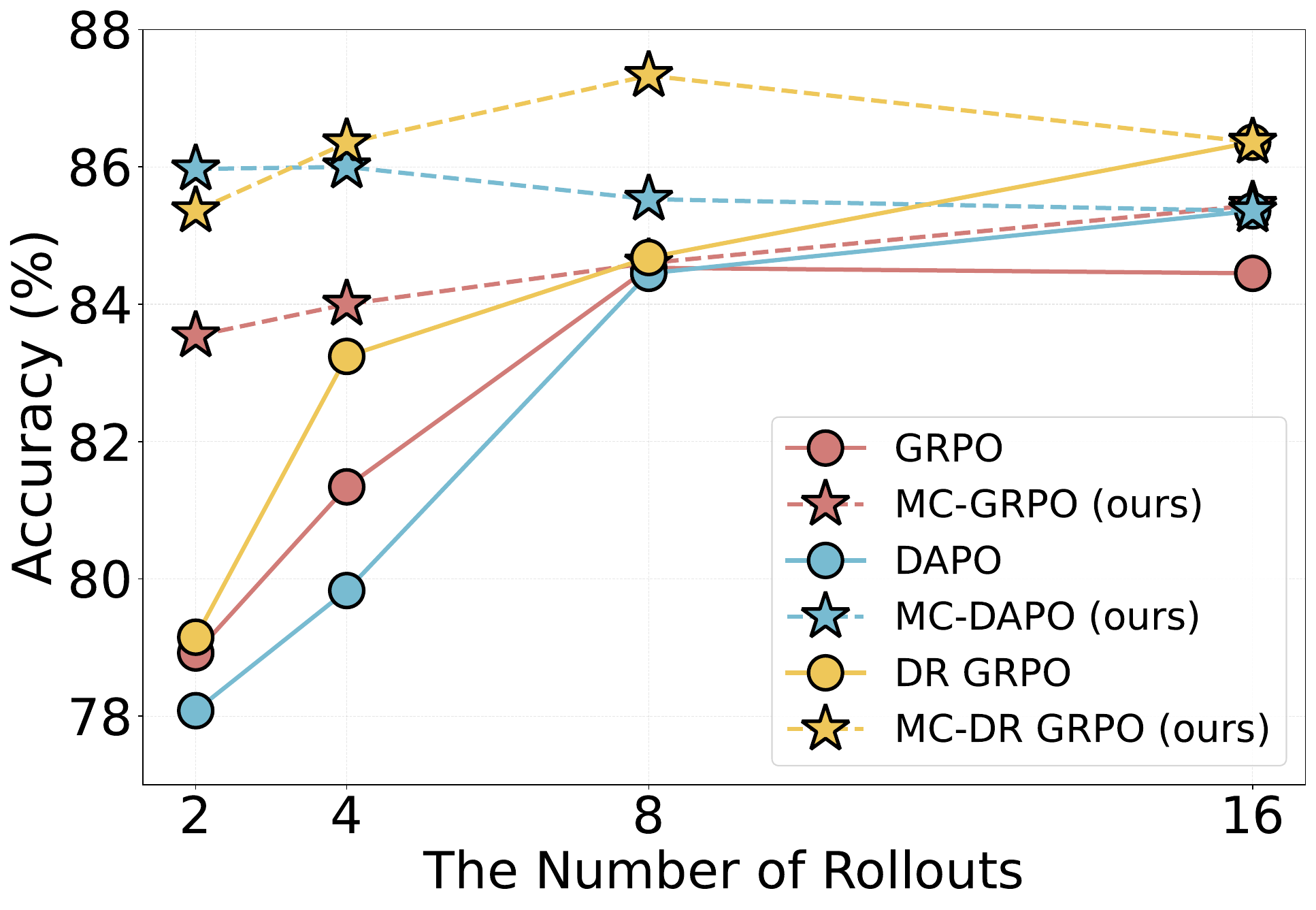}
  \vspace{-1mm}
\caption{
Accuracy (\%) versus the number of rollouts for Qwen3-1.7B trained on GSM8K.
We compare the original GRPO, DAPO, and DR-GRPO methods (\capcirc; baselines) with their Median-Centered (MC) variants (\capstar; ours).
MC training improves robustness and yields larger gains under small rollout budgets (2$\sim$4 rollouts), while remaining competitive at higher rollout counts.}
  \label{fig:rollout_sweep_main}
  \vspace{-3mm}
\end{figure}

Despite its effectiveness at high rollout counts (\eg, 8--32), this design becomes fragile when training is resource-constrained.
In practice, the number of rollouts per prompt is often limited by throughput, latency, or memory constraints~\cite{wu2025takes,lin2025cppo}, especially for individual researchers or academic settings with limited GPU resources.
From a standard RL perspective, reducing the rollout budget naturally hurts performance (Fig. \ref{fig:rollout_sweep_main}) because it weakens exploration: with fewer rollouts, the policy observes fewer diverse candidates and is less likely to sample rare but high-quality trajectories \cite{sutton2018reinforcement}.

In our work, we find that reduced exploration is not the only story behind performance degradation in the small-rollout regime.
In addition, accuracy drops are driven by \emph{advantage sign flips} (Sec.~\ref{sec:motivation}).
Specifically, in GRPO-style methods, the advantage for each rollout is computed relative to the mean reward across rollouts, so higher-reward completions should receive positive advantages while lower-reward ones receive negative advantages (Eq.~\ref{eq:grpo_adv_meanstd}).
When $G$ is small, a single outlier reward can substantially shift this mean baseline, causing multiple rollouts to flip the sign of their advantages.
Such sign flips reverse the intended update direction, \ie, reinforcing worse trajectories and penalizing better ones; this effect compounds over training and leads to lower final performance (even a 5\% sign-flip rate induces a $\sim$4\% accuracy drop).

To address this issue, we propose {Median-Centered Group Relative Policy Optimization (MC-GRPO)}, replaces mean-centering with \emph{median-centering} when computing the shared within-prompt baseline.
The median is a classical robust estimator that is far less sensitive to outliers, and in our setting it directly targets the baseline-induced advantage sign-flip failure mode.
Concretely, we sample \emph{one extra rollout} per prompt to form an odd-sized group $(G{+}1)$.
With an odd-sized group, exactly one completion is the median and therefore receives zero advantage; we exploit this property and \emph{exclude} the median sample from backpropagation, keeping the number of gradient-contributing samples per prompt equal to $G$ and thus preserving the core update cost of the standard $G$-rollout method.
In practice, only one additional rollout often brings negligible cost with high-throughput inference engines (\eg, vLLM).

Importantly, MC training is \emph{algorithm- and model-agnostic} within the GRPO family: applying the same median-centered baseline yields consistent improvements in both training stability and final task performance across GRPO, DAPO \cite{yu2025dapo}, and DR-GRPO \cite{liu2025understanding}. We also observe the same trend across a broad set of models ranging from small to mid-sized LLMs, including Llama-3.2-3B-Instruct \cite{dubey2024llama3}, Qwen2.5-Math-1.5B \cite{qwen25_math_techreport_2024}, Qwen3-1.7B, Qwen3-4B-Instruct \cite{qwen3_techreport_2025}, and Qwen2.5-7B-Instruct \cite{qwen25_techreport_2024}. Our proposed method substantially narrows the performance gap between low- and high-rollout regimes. For Qwen3-1.7B on GSM8K, GRPO achieves 78.90\% with 2 rollouts and 84.53\% with 8 rollouts. With our method, the 2-rollout setting improves to 83.54\%, substantially narrowing the gap to the 8-rollout result.

Our contribution can be summarized as follows:
(1) We identify \emph{baseline-induced advantage sign flips} as a key driver of the accuracy drop in GRPO-style online RL under small rollout budgets. When the shared mean baseline is noisy, reversing update directions and compounding errors over training.
(2) We propose MC-GRPO, a simple and effective method that samples one additional rollout to form an odd-sized group and replaces mean-centering with robust median-centering. The median sample has zero advantage and can be excluded from backprop, preserving the core update cost while directly targeting sign-flip instability.
(3) Across various GRPO-family algorithms and across diverse model families and scales, MC training consistently improves stability and final accuracy in the low-budget regime, achieving $\le 1\%$ gap between $2$-rollout and  $8$-rollout.

\section{Motivation}
\label{sec:motivation}

\begin{figure*}[t]
  \centering
  \includegraphics[width=0.99\linewidth]{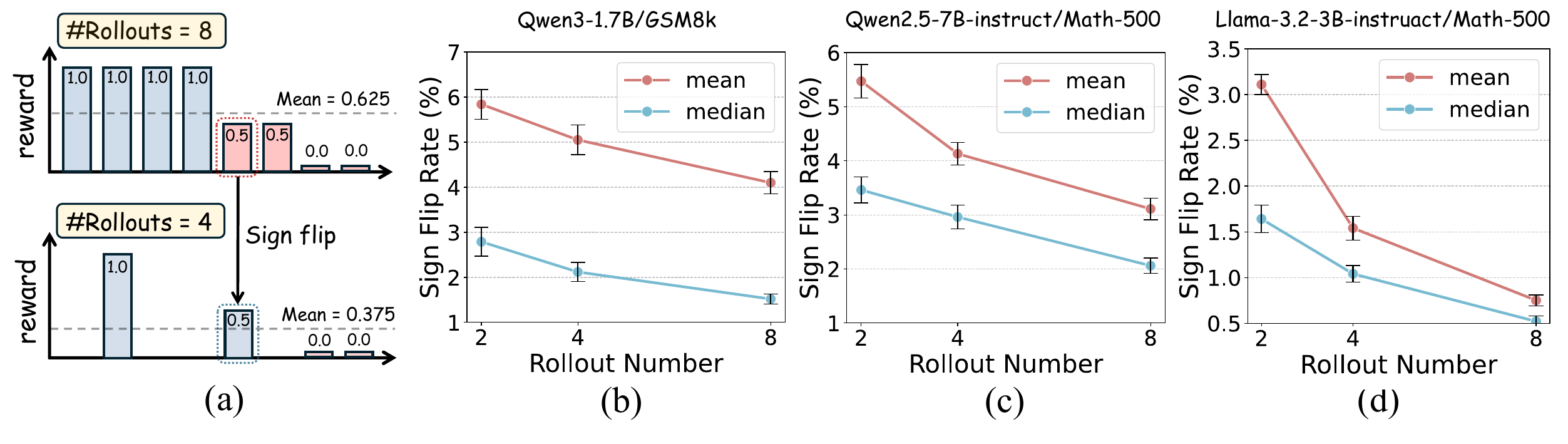}
  \vspace{-1mm}
  \caption{\textbf{Sign flips are frequent under small rollout budgets.}
  (a) With few rollouts, the sample-mean baseline can shift substantially depending on which rollouts are included, causing an advantage sign flip for the same trajectory (\eg, the $0.5$-reward sample flips sign when the rollout set changes from $8$ to $4$).
  We also report the \emph{sign-flip rate} (\ie the fraction of rollouts whose advantage sign under a $k$-rollout baseline disagrees with an oracle sign computed from $G_{\mathrm{ref}}{=}128$ rollouts) for (b) Qwen3-1.7B/GSM8K, (c) Qwen2.5-7B-Instruct/Math-500, and (d) Llama-3.2-3B-instruct/Math-500.
  For each setting, we evaluate 250 prompts; for each prompt and $k\in\{2,4,8\}$, we draw 20 random $k$-subsamples from the 128 rollouts, compute either the mean or median baseline, and average the resulting sign-flip rates.}
  \label{fig:motivation}
\end{figure*}

Group-Relative Policy Optimization (GRPO) and its variants adapt PPO by computing advantages relative to other rollouts from the same prompt \cite{shao2024deepseekmath,yu2025dapo,liu2025understanding}.
Given a prompt $q$, GRPO samples $G$ rollout completions $o_1,\dots,o_G \sim \pi_{\theta_{\mathrm{old}}}(\cdot\mid q)$ and evaluates rewards $r_i = R(q,o_i)$, where $R(\cdot)$ and $\pi_{\theta_{\mathrm{old}}}$ denote the reward function and the policy model, respectively.
It then forms a group-normalized advantage using the within-group mean and standard deviation,
\begin{equation}
\bar r(q)=\frac{1}{G}\sum_{j=1}^{G} r_j,
\qquad
A_i=\frac{r_i-\bar r(q)}{s_r(q)+\varepsilon},
\label{eq:grpo_adv_meanstd}
\end{equation}
and uses $A_i$ in a gradient update.
Crucially, the baseline mean $\bar r(q)$ and standard deviation $s_r(q)$ are {shared across all rollouts} for the same prompt, so any estimation error simultaneously perturbs the advantages for the entire group.

\paragraph{Small-rollout failure mode: baseline-induced sign flips.}
When the rollout count $G$ is small, these shared group statistics become noisy.
In this regime, even a single high-reward rollout can sharply shift the group mean.
Because the baseline is shared, this shift moves the decision boundary for \emph{every} rollout in the group and can flip the sign of some advantages, reversing the update direction for those trajectories.
Fig.~\ref{fig:motivation}(a) illustrates a simple example: the same reward completion switches from positive to negative advantage when the rollout set changes from $8$ (large rollout) to $4$ (small rollout).
We quantify this phenomenon with the \emph{sign-flip rate} (Fig.~\ref{fig:motivation}(b--d)).
Across three different settings (Qwen3-1.7B/GSM8K, Qwen2.5-7B-Instruct/Math-500, and Llama-3.2-3B/Math-500), the sign-flip rate is highest at very small budgets ($k\in\{2,4\}$), confirming that baseline noise is a practical small-rollout issue rather than an artifact of a single model or reward function.

\paragraph{Motivation: robust baselines.}
These observations motivate replacing mean-centering with a more robust location estimator.
The group median is known to be far less sensitive to outliers, stabilizing the shared baseline and reducing advantage sign flips \cite{huber2009robust,rousseeuw2011outlier}.
Consistent with this intuition, median baselines yield systematically lower sign-flip rates than mean baselines across all settings, with the largest improvements at $k\in\{2,4\}$ where mean shifts are most pronounced.
This motivates our \emph{Median-Centered GRPO (MC-GRPO)}.

\begin{figure}[t]
  \centering
  \includegraphics[width=0.99\linewidth]{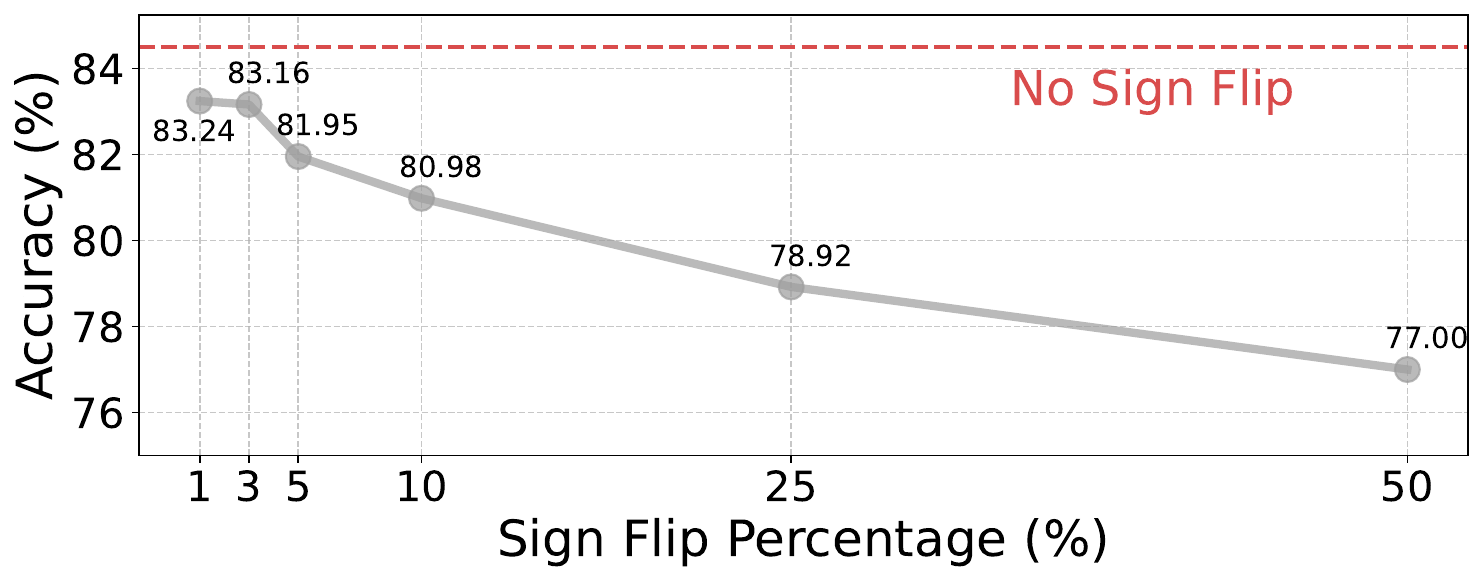}
    \vspace{-1mm}
  \caption{\textbf{Injected sign flips causally degrade GRPO training.}
  Qwen3-1.7B is trained on GSM8K under a fixed rollout budget ($G{=}8$), while we synthetically inject sign noise by flipping the sign of a fraction $\rho$ of within-group advantages during training.
  Increasing $\rho$ consistently reduces final GSM8K accuracy, showing that the advantage sign flips directly corrupt the update direction and harm optimization.}
  \label{fig:sign_flip}
  \vspace{-2mm}
\end{figure}

\paragraph{Sign-flip rate predicts optimization quality.}
A GRPO-style update is ultimately driven by the \emph{direction} of the within-group advantage: trajectories with $A_i>0$ are reinforced while those with $A_i<0$ are suppressed.
When baseline noise flips the sign of $A_i$ for a fixed completion, the update for that trajectory is reversed, \ie, good rollouts can be penalized and bad rollouts can be rewarded.
Because the same noisy baseline is shared across the entire group, a small baseline shift can flip many advantages at once, making sign flips a direct indicator of \emph{optimization direction errors} rather than a benign measurement artifact.

To validate this connection, we run a controlled experiment on {Qwen3-1.7B trained on GSM8K} under a fixed rollout budget.
Starting from standard GRPO with $G{=}8$, we \emph{inject sign noise} by randomly flipping the sign of a fraction $\rho$ of advantages within each prompt group, while keeping all other training settings unchanged.
As $\rho$ increases, final GSM8K accuracy degrades monotonically (Fig.~\ref{fig:sign_flip}), confirming that sign flips causally harm learning by corrupting the update direction.
This makes the sign-flip rate a practical diagnostic: configurations with higher sign-flip rates reliably correspond to poorer downstream performance.

Our MC baselines directly target this failure mode.
By stabilizing the shared location/scale statistics within each prompt group, median-centering substantially reduces sign flips, thereby increasing the fraction of updates applied in the correct direction.
This link explains why MC-GRPO yields the largest gains in the low-budget regime (\eg, $G\in\{2,4\}$) while remaining competitive at larger budgets.

\begin{figure*}[t]
  \centering
  \includegraphics[width=0.90\linewidth]{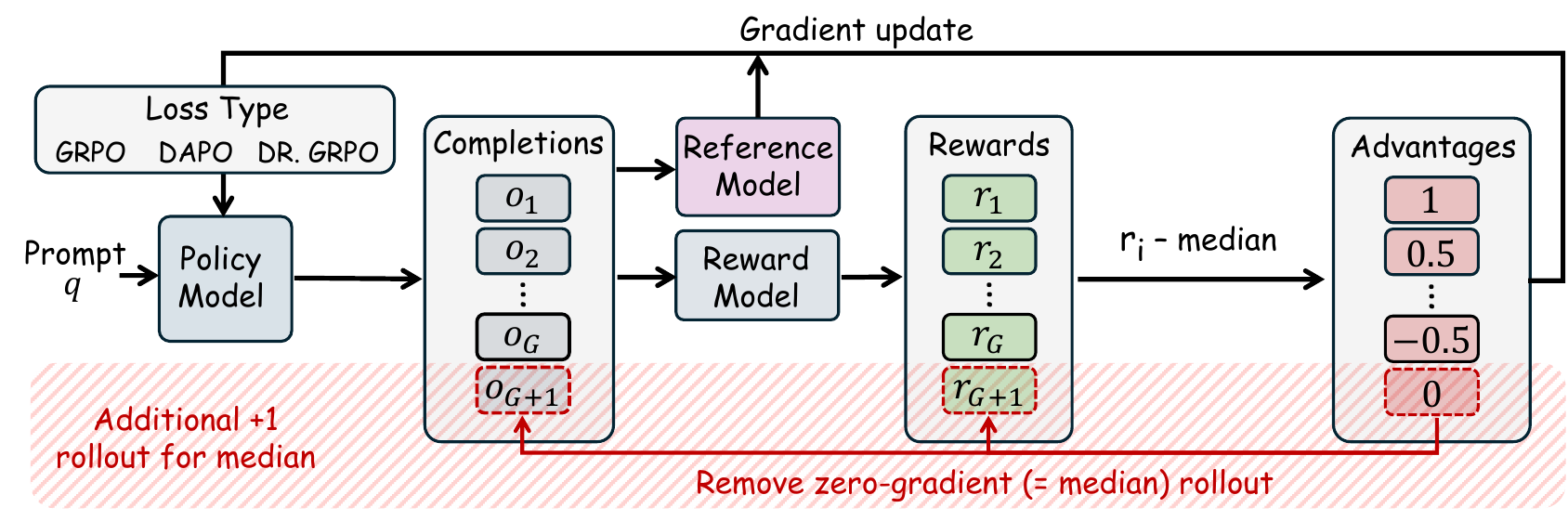}
\caption{{MC-GRPO overview.}
Given a prompt $q$, the policy samples $G{+}1$ completions and obtains rewards via a reward model (and, when applicable, a reference-model term).
We compute group advantages by median-centering around $b(q)=\mathrm{median}(r_1,\dots,r_{G+1})$, which provides a robust shared baseline and reduces sensitivity to occasional high-reward outliers under small rollout budgets.
To keep the effective update size fixed at $G$ trajectories per prompt, we generate one additional completion to define a unique median and then remove the median (zero-advantage) completion from the gradient update.
The resulting advantages are a drop-in replacement for the original group-normalized advantages in standard GRPO-family losses (GRPO/DAPO/DR.GRPO).}
  \label{fig:method}
\end{figure*}

\section{Methodology}
\label{sec:method}

Our motivation suggests that under small rollout budgets, the shared \emph{sample-mean} baseline can be unstable, which may flip advantage signs and degrade update quality (Fig.~\ref{fig:motivation}).
We therefore replace mean-centering in GRPO-style objectives with a robust \emph{median} baseline.
We call the resulting variant \emph{Median-Centered GRPO (MC-GRPO)}.

Following the notation in Eq.~\eqref{eq:grpo_adv_meanstd} (Section~\ref{sec:motivation}), given a prompt $q$, we sample an odd-sized group of $G{+}1$ rollout completions
$o_1,\dots,o_{G+1} \sim \pi_{\theta_{\mathrm{old}}}(\cdot \mid q)$ and evaluate rewards $r_i = R(q,o_i)$.
We define the group baseline as the median
\begin{equation}
b(q) \;=\; \mathrm{median}\!\left(r_1,\dots,r_{G+1}\right),
\label{eq:mcgrpo_median_baseline}
\end{equation}
and compute median-centered, group-normalized advantages
\begin{equation}
A_i \;=\; \frac{r_i - b(q)}{\mathrm{MAD}(r)+\varepsilon},
\qquad i \in \{1,\dots,G{+}1\},
\label{eq:mcgrpo_adv}
\end{equation}
where $\varepsilon>0$ prevents division by zero and the median absolute deviation (MAD) is
\begin{equation}
\mathrm{MAD}(r)
=
\mathrm{median}\!\left(\left|r_1-b(q)\right|,\dots,\left|r_{G+1}-b(q)\right|\right).
\label{eq:mcgrpo_mad}
\end{equation}
We use MAD because it is a standard robust scale estimator. Unlike the sample standard deviation, MAD is far less sensitive to rare high-reward outliers that can dominate within-group normalization under small $G$ \cite{rousseeuw2011outlier}.

\paragraph{Remove the median (zero-gradient) completion.}
Our optimization objective follows the standard GRPO-style clipped surrogate, which broadcasts the sequence-level advantage $A_i$ to token-level updates:
\begin{equation}
\small
\mathcal{J}(\theta)
=
\mathbb{E}
\Bigg[
\frac{1}{G}\sum_{i=1}^{G}\frac{1}{|o_i|}\sum_{t=1}^{|o_i|}
\min\Big(
\rho_{i,t}(\theta)\,A_i,
\hat\rho_{i,t}(\theta)\,A_i
\Big)
\Bigg],
\label{eq:grpo_obj}
\end{equation}
where $o_{i,t}$ is the $t$-th token in completion $o_i$,
$\rho_{i,t}(\theta)=\pi_{\theta}(o_{i,t}\mid q,o_{i,<t})/\pi_{\theta_{\mathrm{old}}}(o_{i,t}\mid q,o_{i,<t})$, and
$\hat\rho_{i,t}(\theta)=\mathrm{clip} \big(\rho_{i,t}(\theta), 1-\epsilon, 1+\epsilon\big)$.

In our MC-GRPO, we sample one additional rollout, yielding $G{+}1$ completions.
Let $i^\star$ denote the index such that $r_{i^\star}=b(q)$, then $A_{i^\star}=0$ by Eq.~\eqref{eq:mcgrpo_adv}.
To keep the effective update size fixed at $G$ trajectories per prompt, we exclude the median completion:
\begin{equation}
\mathcal{I}(q) \;=\; \{1,\dots,G+1\}\setminus\{i^\star\}.
\qquad |\mathcal{I}(q)|=G.
\label{eq:mcgrpo_indexset}
\end{equation}

The rationale for excluding the median completion (with $A_{i^\star}=0$) from the GRPO policy gradient is that it contributes trivially to the update.
Since the gradient is additive over rollout completions, we can separate the median index $i^\star$ from the remaining non-median set $\mathcal{I}(q)$ and write 
\begin{equation} \small \hspace{-10mm}\begin{aligned} \nabla_\theta \mathcal{J}(\theta) \hspace{-1mm} \;\approx\; \hspace{-1mm} \mathbb{E}\Bigg[ \frac{1}{G}\Bigg( \sum_{i\in\mathcal{I}(q)}\frac{1}{|o_i|}\sum_{t=1}^{|o_i|} \rho_{i,t}(\theta)\,A_i\, \nabla_\theta \log \pi_\theta\!\big(o_{i,t}\mid q,o_{i,<t}\big) \\ \qquad\qquad\qquad\qquad +\; \frac{1}{|o_{i^\star}|}\sum_{t=1}^{|o_{i^\star}|}\rho_{i^\star,t}(\theta)\,A_{i^\star}\, \nabla_\theta \log \pi_\theta\!\big(o_{i^\star,t}\mid q,o_{i^\star,<t}\big) \Bigg) \Bigg]. \end{aligned} \label{eq:grpo_grad_decomp_split} \end{equation}
The median completion satisfies $A_{i^\star}=0$, so the second term vanishes.
Therefore, excluding $i^\star$ leaves the gradient estimate unchanged while keeping the effective update size fixed at $G$.
For clarity, we omit the KL-regularization term here, as it typically has a small effect on the gradient in our setting. We provide details in Appendix C.

\paragraph{Plug-in objective for GRPO variants.}
MC-GRPO only changes how group advantages are computed: we replace the mean-centered baseline with a median-centered baseline, while leaving the underlying GRPO-family objective and optimization pipeline unchanged.
Concretely, for any GRPO-style clipped surrogate, we simply substitute its original group-normalized advantages with our median-centered $A_i$ from Eq.~\eqref{eq:mcgrpo_adv}, and keep all other components identical (\eg, the importance ratio $\rho_{i,t}$, clipping with $\epsilon$, KL regularization, and any variant-specific loss terms).
Because this change is confined to the advantage estimator, MC-GRPO is directly compatible with GRPO variants such as DAPO and DR-GRPO: it can be applied by replacing their within-group advantage computation with Eq.~\eqref{eq:mcgrpo_adv}, without modifying any other part of the objective.
We summarize the overall procedure in Algorithm~\ref{alg:mcgrpo}.

\begin{algorithm}[t]
\caption{Median-Centered GRPO (MC-GRPO)}
\label{alg:mcgrpo}
\begin{algorithmic}[1]
\STATE \textbf{Input:} prompt batch $\mathcal{Q}$, old policy $\pi_{\theta_{\mathrm{old}}}$, current policy $\pi_\theta$, rollout budget $G$, clip $\epsilon$, small constant $\varepsilon>0$
\FOR{each $q \in \mathcal{Q}$}
  \STATE Generate $G{+}1$ rollout $o_1,\dots,o_{G+1} \sim \pi_{\theta_{\mathrm{old}}}(\cdot\mid q)$
  \STATE Compute rewards $r_i \leftarrow R(q,o_i)$ for $i=1,\dots,G{+}1$
  \STATE $b(q) \leftarrow \mathrm{median}(r_1,\dots,r_{G+1})$ 
  \STATE $\mathrm{MAD}(r) \leftarrow \mathrm{median}(|r_1-b(q)|,\dots,|r_{G+1}-b(q)|)$ 
  \FOR{$i=1$ to $G{+}1$}
    \STATE $A_i \leftarrow (r_i-b(q))/(\mathrm{MAD}(r)+\varepsilon)$ \hfill (Eq.~\eqref{eq:mcgrpo_adv})
  \ENDFOR
  \STATE $\mathcal{I}(q) \leftarrow \{1,\dots,G{+}1\}\setminus\{i^\star\}$ \hfill (Eq.~\eqref{eq:mcgrpo_indexset})
  \STATE Optimize a GRPO objective for non-zero  $A_i$ completion (Eq.~\eqref{eq:grpo_obj}) 
\ENDFOR
\end{algorithmic}
\end{algorithm}

\section{Experiments}
\label{sec:experiments}

\subsection{Experimental Settings}
\label{sec:exp_settings}

\paragraph{Models, Datasets, and Training details.}
We evaluate MC-GRPO on two math-reasoning tasks across five model--dataset configurations: GSM8K with Qwen3-1.7B and Llama-3.2-3B, and Math-500 with Qwen2.5-Math-1.5B, Qwen3-4B-Instruct, and Qwen2.5-7B-Instruct. For GSM8K, we run RL training on the official training split and report exact-match accuracy on the standard test split~\cite{cobbe2021gsm8k}. For Math-500, we use the MATH dataset~\cite{hendrycks2021math} with LightEval-compatible preprocessing from \texttt{DigitalLearningGmbH/MATH-lighteval}, and evaluate on the \texttt{math500} split (a 500-problem subset). We further evaluate on AMC 2023~\cite{maa_amc2023} and AIME 2024~\cite{maa_aime2024} to assess generalization to competition-style math reasoning. Training details are provided in Appendix~B.

\paragraph{Baselines.}
We compare MC-GRPO against its corresponding GRPO-family baseline, which computes group advantages by mean-centering and mean/std normalization, while matching the rollout budget and the overall training recipe.
For GRPO variants such as DAPO or DR-GRPO, we keep their original objective (as implemented in TRL) unchanged and only replace the advantage estimator with our median-centered advantages.

\begin{table*}[t]
  \centering
\caption{{Results under different GRPO rollout budgets.}
We report final accuracy and wall-clock training time as a function of the rollout budget $G$ across five model--dataset settings (GSM8K and Math-500).
MC-GRPO constructs a median baseline using additional sampled completions and removes the median (zero-advantage) completion so that the number of completions contributing to the policy-gradient remains $G$, matching GRPO.
Green numbers indicate absolute accuracy improvements over GRPO at the same number of training samples per prompt.}
\label{tab:main_results}
  \vspace{-1mm}
  \small
  \setlength{\tabcolsep}{4pt}
  \renewcommand{\arraystretch}{1.05}

  \resizebox{0.95\textwidth}{!}{%
  \begin{tabular}{l| c |l c c l c}
    \toprule
    Model & Dataset & Method & \#Rollout ($G$) & \#Training samples & Accuracy (\%) & Training time (sec) \\
     \midrule

    \multirow{6}{*}{Qwen3-1.7B} & \multirow{6}{*}{GSM8K}
      & GRPO & 8 & 8 & 84.53 & 4739 \\
    & & \mcshade{MC-GRPO (ours)} & \mcshade{8+1} & \mcshade{8} & \mcshade{84.60~\gain{(+0.07)}} & \mcshade{5054} \\
    & & GRPO & 4  & 4 & 81.34 & 2383 \\
    & & \mcshade{MC-GRPO (ours)} & \mcshade{4+1} & \mcshade{4} & \mcshade{84.01~\gain{(+2.67)}} & \mcshade{2561} \\
    & & GRPO & 2  & 2 & 78.92 &  1316  \\
    & & \mcshade{MC-GRPO (ours)} & \mcshade{3+1} & \mcshade{2} & \mcshade{83.54~\gain{(+4.62)}} & \mcshade{1521}  \\
     \midrule

    \multirow{6}{*}{Llama-3.2-3B-Instruct} & \multirow{6}{*}{GSM8K}
      & GRPO & 8 & 8 & 79.00 & 6341 \\
    & & \mcshade{MC-GRPO (ours)} & \mcshade{8+1} & \mcshade{8} & \mcshade{79.02~\gain{(+0.02)}} & \mcshade{6894} \\
    & & GRPO & 4  & 4 & 77.33 & 2930 \\
    & & \mcshade{MC-GRPO (ours)} & \mcshade{4+1} & \mcshade{4} & \mcshade{79.68~\gain{(+2.35)}} & \mcshade{3298} \\
    & & GRPO & 2  & 2 & 76.57 & 1606 \\
    & & \mcshade{MC-GRPO (ours)} & \mcshade{3+1} & \mcshade{2} & \mcshade{{77.93~\gain{(+1.36)}}}
    & \mcshade{1945} \\     \midrule

    \multirow{6}{*}{Qwen2.5-Math-1.5B} & \multirow{6}{*}{Math-500}
      & GRPO & 8 & 8 & 70.00 & 13774 \\
    & & \mcshade{MC-GRPO (ours)} & \mcshade{8+1} & \mcshade{8} & \mcshade{70.60~\gain{(+0.60)}} & \mcshade{14585} \\
    & & GRPO & 4  & 4 & 67.80 & 6990 \\
    & & \mcshade{MC-GRPO (ours)} & \mcshade{4+1} & \mcshade{4} & \mcshade{70.15~\gain{(+2.35)}} & \mcshade{7674} \\
    & & GRPO & 2  & 2 & 65.01 & 3406 \\
    & & \mcshade{MC-GRPO (ours)} & \mcshade{2+1} & \mcshade{2} & \mcshade{68.62~\gain{(+3.61)}} & \mcshade{3923} \\
     \midrule

    \multirow{6}{*}{Qwen3-4B-Instruct} & \multirow{6}{*}{Math-500}
      & GRPO & 8 & 8 & 80.21 & 20841 \\
    & & \mcshade{MC-GRPO (ours)} & \mcshade{8+1} & \mcshade{8} & \mcshade{82.83~\gain{(+2.62)}} & \mcshade{22728} \\
    & & GRPO & 4  & 4 & 80.21 & 11838 \\
    & & \mcshade{MC-GRPO (ours)} & \mcshade{4+1} & \mcshade{4} & \mcshade{82.80~\gain{(+2.59)}} & \mcshade{13336} \\
    & & GRPO & 2  & 2 & 78.80 & 6250  \\
    & & \mcshade{MC-GRPO (ours)} & \mcshade{2+1} & \mcshade{2} & \mcshade{81.92~\gain{(+3.12)}} & \mcshade{7150} \\
     \midrule

    \multirow{6}{*}{Qwen2.5-7B-Instruct} & \multirow{6}{*}{Math-500}
      & GRPO & 8 & 8 & 75.8 & 21949 \\
    & & \mcshade{MC-GRPO (ours)} & \mcshade{8+1} & \mcshade{8} & \mcshade{77.0~\gain{(+1.2)}} & \mcshade{27700} \\
    & & GRPO & 4  & 4 & 74.4 & 11867 \\
    & & \mcshade{MC-GRPO (ours)} & \mcshade{4+1} & \mcshade{4} & \mcshade{76.6~\gain{(+2.2)}} & \mcshade{14427} \\
    & & GRPO & 2  & 2 & 72.2 & 6498 \\
    & & \mcshade{MC-GRPO (ours)} & \mcshade{2+1} & \mcshade{2} & \mcshade{76.8~\gain{(+4.6)}} & \mcshade{7976} \\

    \bottomrule
  \end{tabular}%
  }
  \vspace{-2mm}
\end{table*}

\begin{figure*}[t]
  \centering
  \includegraphics[width=1.0\linewidth]{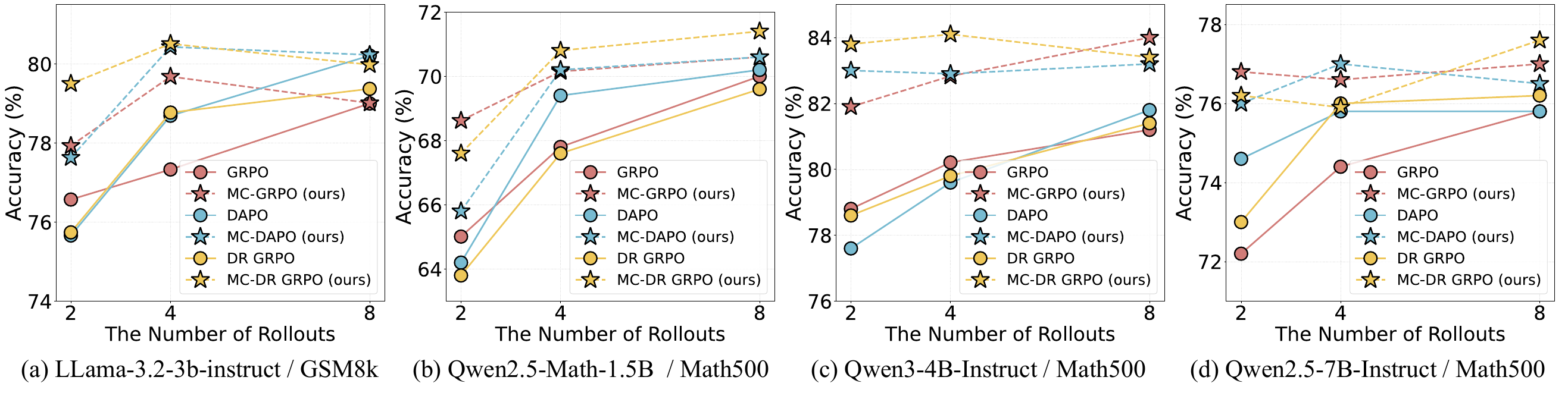}
\caption{\textbf{MC-GRPO overview.}
Accuracy as a function of the rollout budget $G\in\{2,4,8\}$ for GRPO, DAPO, and DR-GRPO, and their median-centered variants (MC-GRPO, MC-DAPO, MC-DR-GRPO).
Across all settings, median-centering yields the largest gains at small rollout budgets and remains competitive as $G$ increases.
}
  \label{fig:overall_accuracy}
\end{figure*}

\begin{figure}[t]
  \centering
  \begin{subfigure}[t]{0.495\linewidth}
    \centering
    \includegraphics[width=\linewidth]{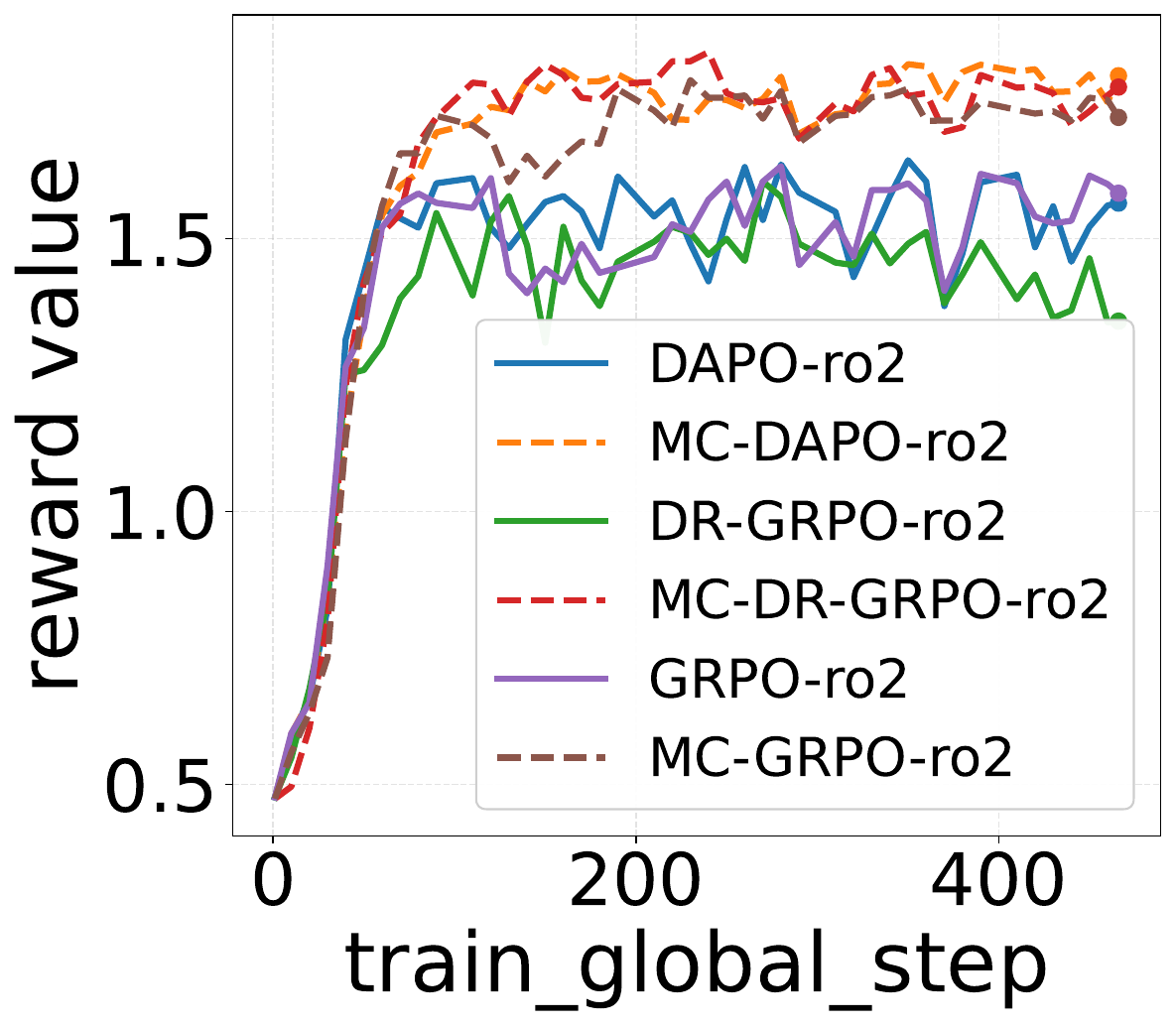}
  \end{subfigure}
  \hfill
  \begin{subfigure}[t]{0.495\linewidth}
    \centering
    \includegraphics[width=\linewidth]{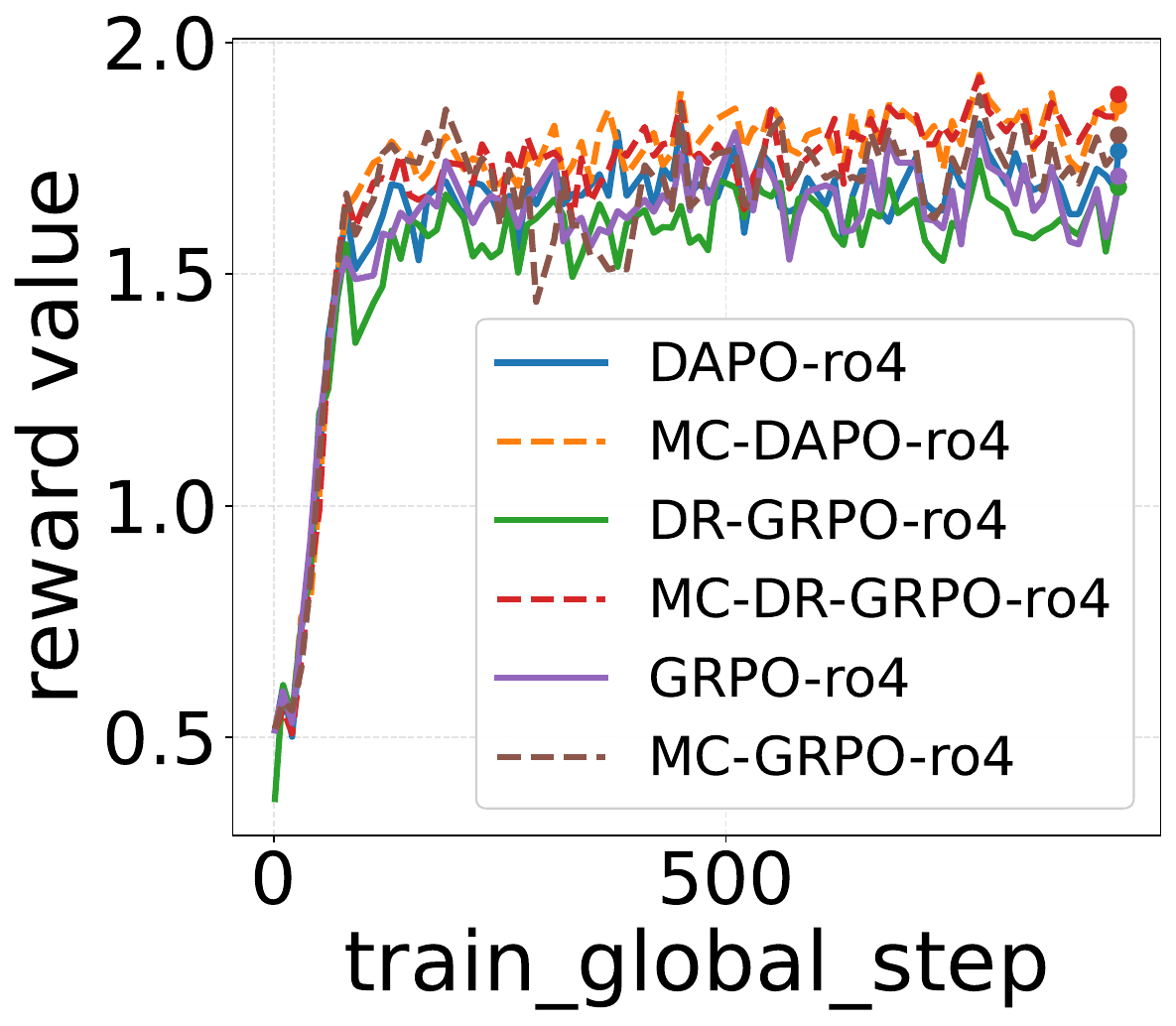}
  \end{subfigure}
\caption{{Training reward dynamics under small rollout budgets.}
({Left}) With rollout $G{=}2$, the median-centered variants  consistently achieve higher and more stable training reward than their mean-centered counterparts, matching our motivation that robust baselines reduce baseline-induced noise and advantage sign flips.
({Right}) With rollout $G{=}4$, the gap between MC and the original GRPO-family methods narrows as the mean baseline becomes more reliable, but MC variants still maintain a modest reward advantage across training steps. The experiments are based on  LLama-3.2-3b-instruct / GSM8k setting.}
  \label{fig:train_reward}
\end{figure}

\subsection{Small-rollout performance}
\label{sec:exp_small_rollout}

Table~\ref{tab:main_results} evaluates the impact of the rollout budget $G$ on accuracy and training time.
Across all model--dataset settings, MC-GRPO improves performance in the small-rollout regime ($G\in{2,4}$), where GRPO-style updates are sensitive to baseline noise and advantage sign flips (Sec.~\ref{sec:motivation}).
The gains are largest at small rollout budgets, reaching up to a $4.62\%$ improvement on GSM8K and a $4.6\%$ improvement on Math-500 at $G=2$, and yielding $2.35\%\ \sim 2.67\%$ improvements at $G=4$ across models and datasets.
As the rollout budget increases, the gains from MC-GRPO diminish, which is expected because the mean baseline is estimated more accurately with more rollouts.
MC-GRPO samples additional completions to define a robust median baseline while keeping the effective update size matched to GRPO.
Specifically, for each prompt, we retain exactly $G$ completions in the policy gradient by discarding the median completion with zero advantage.
The extra rollouts increase wall-clock training time due to additional generation, while the number of training samples contributing to the gradient remains fixed.

We also use the same median-centered advantage estimator as a  replacement for the advantage computation in other GRPO-family objectives, keeping their original loss terms unchanged.
As shown in Fig.~\ref{fig:overall_accuracy}, the median-centered variants of DAPO and DR-GRPO follow the same trend as MC-GRPO: improvements are most pronounced when the rollout budget is small ($G\in\{2,4\}$), and the benefit diminishes as $G$ increases while remaining competitive at $G=8$.

\subsection{Analysis}

\paragraph{Training reward dynamics.}
Fig.~\ref{fig:train_reward} plots the average training reward over optimization steps under small rollout budgets.
Across GRPO-family methods, the median-centered variants (MC-GRPO, MC-DAPO, and MC-DR-GRPO) improve more quickly in the early stage and converge to higher reward levels with noticeably reduced fluctuations compared to their mean-centered counterparts.
The effect is strongest at $G{=}2$, where mean-based baselines are most susceptible to outlier rollouts and discretized reward noise, resulting in noisy advantage estimates and less stable updates.
At $G{=}4$, the gap decreases as the mean baseline becomes better estimated, but the median-centered variants still maintain a modest reward advantage throughout training.
These dynamics are consistent with our hypothesis that median-centered baselines stabilize within-prompt advantage estimation and mitigate update noise, leading to more reliable reward acquisition during training.

\begin{figure}[t]
  \centering
  \includegraphics[width=\linewidth]{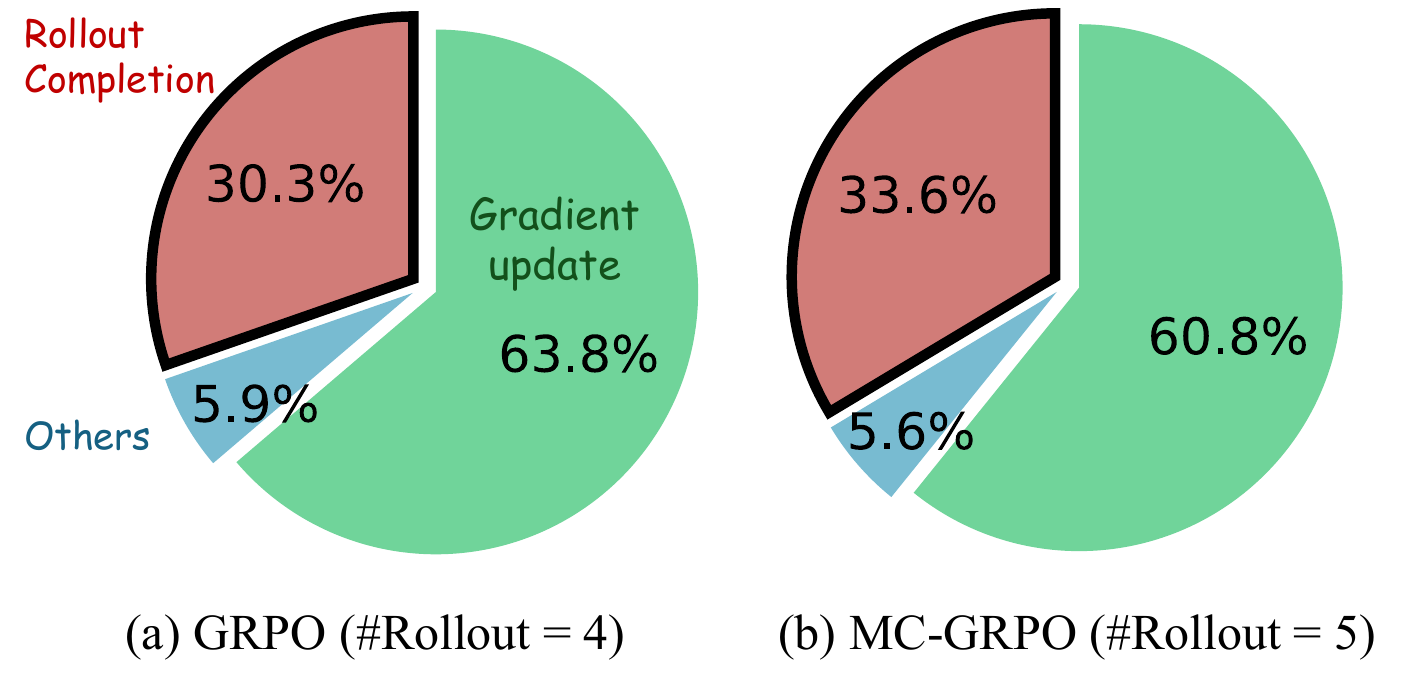}
  \caption{
Wall-clock breakdown of GRPO (left, $G{=}4$) and MC-GRPO (right, $G{+}1{=}5$).
In both cases, gradient updates dominate the total runtime, while rollout completion and other components account for a smaller fraction.
Sampling one additional rollout to form the median baseline therefore increases end-to-end training time only marginally.}
\label{fig:compute_breakdown}
\end{figure}

\paragraph{Latency breakdown under small rollout budgets.}
A natural concern is that MC-GRPO may benefit simply from sampling more rollouts, since it draws one additional completion to form an odd-sized group and define a median baseline.
Fig.~\ref{fig:compute_breakdown} addresses this concern by showing a wall-clock breakdown of GRPO ($G{=}4$) and MC-GRPO ($G{+}1{=}5$).
In both methods, the dominant cost comes from gradient updates, while rollout completion and other components constitute a smaller portion of the end-to-end pipeline.
As a result, adding a single rollout introduces only a modest increase in total training time.
To isolate the effect of baseline robustness from increased sampling, we use an update-size matched protocol throughout: MC-GRPO generates $G{+}1$ rollouts to define the median baseline, then removes the median (zero-advantage) completion so that exactly $G$ samples contribute to the policy-gradient, matching GRPO.
Empirically, MC-GRPO provides the largest gains at small rollout budgets and remains competitive as $G$ increases, and the same trend holds when median-centered advantages are used as a plug-in for other GRPO-family objectives.

\paragraph{Out-of-distribution generalization.}
\label{sec:exp_ood}

We study whether the robustness benefits of median-centered training transfer beyond the training distribution.
We train policies on {Math-500} and evaluate them zero-shot on two harder out-of-distribution contest benchmarks, {AIME-24} and {AMC-23}, using identical test-time prompting and decoding across methods.
We vary only the rollout budget $G$ used during RL training.
Table~\ref{tab:ood_aime_amc} shows that MC-GRPO improves OOD performance in the small-rollout regime.
With $G\in\{2,4\}$, MC-GRPO consistently achieves higher accuracy than mean-centered GRPO on both AIME-24 and AMC-23.
These results mirror the in-distribution trends and are consistent with the mechanism in Section~\ref{sec:motivation}: when $G$ is small, baseline estimation noise can induce advantage sign flips and destabilize the update direction.
Median-centering stabilizes the shared baseline, reducing such direction errors and yielding more reliable learning signals, which translates into better OOD generalization.

\begin{table}[t]
  \centering
 \caption{{OOD generalization on AIME-24 and AMC-23 after training on Math-500.}
We train with GRPO or MC-GRPO on Math-500 using rollout budgets $G\in\{2,4\}$ and report zero-shot accuracy (\%) on the out-of-distribution contest benchmarks AIME-24 and AMC-23.
MC-GRPO consistently improves OOD accuracy under small rollout budgets.}
\label{tab:ood_aime_amc}
\small
  \setlength{\tabcolsep}{6pt}
  \begin{tabular}{l|c|cc}
    \toprule
    Method & \#Rollout ($G$) & AIME-24 & AMC-23 \\
    \midrule
    GRPO & 2 & 16.66 & 67.50\\
    \mcshade{MC-GRPO (ours)} & \mcshade{2+1} & \mcshade{23.33} & \mcshade{72.50} \\
    GRPO & 4 & 13.33 & 67.50 \\
    \mcshade{MC-GRPO (ours)} & \mcshade{4+1} & \mcshade{20.00} & \mcshade{72.50} \\
    \bottomrule
  \end{tabular}
\end{table}

\paragraph{Fine-grained reward robustness: combined discrete rewards.}
Our main GSM8K experiments optimize a partial-credit accuracy reward $r_{\text{acc}}\in\{0,1,2\}$.
In many LLM-RL pipelines, however, rewards can consist of multiple terms, yielding a more fine-grained yet still discrete training signal.
To evaluate robustness in this setting, we augment the accuracy reward with a binary format term $r_{\text{fmt}}\in\{0,1\}$ and optimize the mixed reward
$r = r_{\text{acc}} + r_{\text{format}}$ on {Qwen3-1.7B-Instruct/GSM8K}.
Table~\ref{tab:gsm8k_format_reward} shows that MC-GRPO remains effective under combined rewards.
Across rollout budgets, MC-GRPO consistently outperforms mean-centered GRPO in final GSM8K accuracy, with larger gains at smaller budgets.
This trend is consistent with our mechanism: when $G$ is small, mean-centered baselines are more affected by discrete reward noise and ties, whereas median-centering yields a more stable shared baseline and hence more reliable advantage estimates.

\begin{table}[t]
  \centering
  \caption{{MC-GRPO with composite (fine-grained) rewards on GSM8K.}
  Qwen3-1.7B is trained on GSM8K with a mixed reward
  $r = r_{\text{acc}} + r_{\text{fmt}}$,
  where $r_{\text{acc}}\in\{0,1,2\}$ is the partial-credit accuracy reward and
  $r_{\text{fmt}}\in\{0,1\}$ is a format-compliance reward.
  We report final GSM8K accuracy (\%).
  All numbers are placeholders and will be replaced with actual results.}
  \label{tab:gsm8k_format_reward}
  \small
  \setlength{\tabcolsep}{6pt}
  \begin{tabular}{l|c|l}
    \toprule
    Method & \#Rollout ($G$) & Accuracy (\%) \\
    \midrule
    GRPO & 2 & 79.00\\
    \mcshade{MC-GRPO (ours)} & \mcshade{2+1} & \mcshade{84.14~\gain{(+5.14)}} \\
    GRPO & 4 & 81.34 \\
    \mcshade{MC-GRPO (ours)} & \mcshade{4+1} & \mcshade{84.23~\gain{(+2.89)}} \\
    GRPO & 8 & 84.02 \\
    \mcshade{MC-GRPO (ours)} & \mcshade{8+1} & \mcshade{87.01~\gain{(+2.99)}} \\
    \bottomrule
  \end{tabular}
\end{table}

\paragraph{Disentangling median-centering from extra sampling.}
A natural concern is that MC-GRPO may improve performance simply by drawing one additional rollout, rather than due to the median baseline itself. To isolate the effect of median-centering from extra sampling, we evaluate an update-size matched control on Qwen3-1.7B-Instruct/GSM8K. Specifically, we sample $G{+}1$ rollouts, compute advantages using the standard mean-centered baseline over the $G{+}1$ rewards, and then drop the rollout with the smallest absolute advantage so that exactly $G$ samples contribute to the policy gradient. This control matches MC-GRPO in the number of sampled completions while leaving the baseline estimator unchanged, directly testing whether the observed gains can be attributed to extra sampling alone.

\begin{table}[t]
  \centering
  \caption{
We additionally evaluate an update-size matched {extra-sampling mean} control: we sample $G{+}1$ rollouts, compute mean-centered advantages over the $G{+}1$ rewards, and then discard the rollout with the smallest advantage magnitude so that exactly $G$ samples contribute to the policy gradient. We denote this variant as \texttt{GRPO + 1 (drop one rollout)}. We compare it against MC-GRPO, which uses a median baseline and removes the zero-advantage median completion. The results show that the accuracy gains are not explained by extra rollout number.}

  \label{tab:extra_sampling_control}
  \small
  \setlength{\tabcolsep}{6pt}
  \begin{tabular}{l|c|c}
    \toprule
    Method & \#Rollout ($G$) & Accuracy (\%) \\
    \midrule
    GRPO & 2 & 78.92 \\
    GRPO + 1 (drop one rollout) & 2+1 &  79.68 \\
    \mcshade{MC-GRPO (ours)} & \mcshade{2+1} & \mcshade{83.54} \\
    \midrule
    GRPO & 4 & 81.34 \\
    GRPO + 1 (drop one rollout) & 4+1 & 80.36 \\
    \mcshade{MC-GRPO (ours)} & \mcshade{4+1} & \mcshade{84.01} \\
    \midrule
    GRPO & 8 & 84.53 \\
    GRPO + 1 (drop one rollout) & 8+1 & 83.25\\
    \mcshade{MC-GRPO (ours)} & \mcshade{8+1} & \mcshade{84.60} \\
    \bottomrule
  \end{tabular}
\end{table}
  
\section{Related Work}
\label{sec:related_wrok}

\paragraph{RL for Reasoning LLMs.}
Recent advances in reasoning LLMs combine process supervision and RL-based optimization, building on classic connections between policy gradients and entropy-regularized RL \cite{schulman2017equivalence}. Process-level supervision trains verifiers to validate intermediate steps for more reliable reasoning \cite{lightman2023let}, while the RLHF pipeline operationalized large-scale preference-driven policy optimization \cite{ouyang2022training}. Strong domain-specialized backbones for code and math (\eg, DeepSeek-Coder/DeepSeekMath) provide the foundation on which RL can further amplify long-form reasoning behaviors, exemplified by DeepSeek-R1 and follow-up analyses of R1-zero-like training dynamics \cite{guo2024deepseek,shao2024deepseekmath,guo2025deepseek,liu2025understanding}. On the algorithm/system side, group/sequence-level policy optimization and its stabilized variants (\eg, GSPO, SAPO), along with improved normalization (BNPO), aim to make large-scale reasoning RL more stable and efficient, supported by scalable open-source training stacks like DAPO \cite{zheng2025group,gao2025soft,xiao2025bnpo,yu2025dapo}.

\paragraph{Reasoning Model Acceleration.}
Recent work on accelerating reasoning LLMs spans both inference- and training-time efficiency. On the inference side, methods either \emph{compress/prune} chain-of-thought to reduce decoding cost—via token skipping or shorter CoT generation \cite{xia2025tokenskip,kang2025c3ot}, length-controllable CoT tuning \cite{ma2025cot}, or surprisal-based pruning of predictable steps \cite{zeng2025pruning}—or \emph{adaptively switch} between fast and slow thinking by allocating computation based on difficulty \cite{shen2025dast}, learning policies to trigger CoT under a cost--accuracy trade-off \cite{lou2025adacot}, or routing queries to different reasoning modes \cite{liang2025thinkswitcher}. 
Multiple surveys have recently emerged to systematize these efficiency-oriented reasoning directions \cite{sui2025stop,chen2025towards}.
In parallel, training-time work improves RL efficiency by reducing wasted compute in sampling and updates, e.g., pruning low-signal completions and dynamically allocating rollouts in GRPO-style training (CPPO) \cite{lin2025cppo}, stabilizing critic-free optimization via global normalization (REINFORCE++) \cite{hu2501reinforce++}, and accelerating convergence through online curriculum learning (SPEED-RL) \cite{zhang2025speed}. In contrast, our method targets the \emph{baseline estimator} under small rollout budgets, replacing mean-centering with median-centering to make advantages robust to noise/outliers and reduce flip-prone updates when $G$ is small; it is orthogonal to pruning, normalization, and curriculum selection, and thus naturally compatible with them. 
\section{Conclusion}
\label{sec:conclusion}

This paper revisits a basic but often overlooked source of instability in GRPO-style RL: when only a few rollouts are available per prompt, the \emph{shared} group statistics used for advantage normalization can become unreliable, distorting the learning signal for the entire group. We propose {MC-GRPO}, which replaces mean-centering with a median baseline and keeps the effective update size fixed by dropping the zero-advantage median completion. Despite its simplicity, MC-GRPO consistently improves training reward and final accuracy in the small-rollout regime across multiple model families and math benchmarks, and can be used as a drop-in advantage replacement for GRPO variants, including DAPO and DR-GRPO.
On the other hand, our empirical study is limited to verifier-based, single-objective math rewards on a small set of benchmarks. Such rewards are relatively structured and often close to binary correctness, and thus may exhibit different noise and bias properties than learned reward models or human-preference signals, which can be stochastic, non-stationary, and sensitive to prompt phrasing or stylistic cues. As a result, it remains unclear whether median-centering will deliver the same stability gains in settings with noisier or multi-objective rewards, where a robust within-prompt baseline may need to be task-adaptive or incorporate coordination across objectives.


\nocite{langley00}

\bibliography{example_paper}
\bibliographystyle{icml2026}

\appendix
\onecolumn

\section{GRPO, DAPO, DR-GRPO implementation details}
\label{app:impl_details}

We follow the TRL implementations of GRPO, DAPO, and DR-GRPO and keep their original objectives and training logic unchanged, except where explicitly noted. For DAPO, we omit the Soft Overlong Punishment term, a length-aware penalty designed to shape rewards for truncated/overlong generations when the response length exceeds a predefined maximum, since reward-function design is discussed separately. For DR-GRPO, we do not applying advantage normalization and keep this setting throughout for consistency.

\section{Training details}
We train all models with AdamW~\cite{loshchilov2019adamw} using a learning rate of $1\times10^{-6}$, a cosine learning-rate schedule~\cite{loshchilov2016sgdr}, and a warmup ratio of $0.1$.
per-device batch size 16, and gradient accumulation steps of 1.
We set the KL penalty coefficient to $\beta{=}0.04$.
We run evaluation every 100 (or 500) training steps depending on the dataset.
For rollout generation, we decode with temperature 1.0.
We use an accuracy reward following~\cite{liu2025understanding}.
We use a maximum completion length of 3072 tokens for Math-500.
We sample $G{=}16$ rollouts per prompt for policy optimization; for our median-centered variant, we draw one additional rollout to form an odd-sized group, drop the median completion, and backpropagate through the remaining $G$ samples to keep the effective update size fixed.
We report task accuracy on GSM8K and Math-500 as the primary metric and monitor format accuracy during training.
We log training curves and evaluation metrics with Weights \& Biases.
We implement MC-GRPO in the TRL training pipeline~\cite{trl} and use vLLM for high-throughput colocated rollout generation.
All experiments are run on two NVIDIA B200 GPUs.

\section{Reward Function Design}
\label{app:reward_design}

\paragraph{GSM8K: partial-credit accuracy reward.}
For GSM8K, we use a discrete, partial-credit accuracy reward that assigns higher reward to exact matches while providing intermediate credit for numerically equivalent answers with different surface forms.
Given a model completion, we extract the final answer span using the same post-processing used in evaluation and canonicalize the ground-truth answer accordingly.
The accuracy reward is defined as
\begin{equation}
r_{\text{acc}} =
\begin{cases}
2.0, & \text{if the extracted answer exactly matches the target string},\\
1.5, & \text{if the extracted answer is numerically equivalent to the target},\\
0.0, & \text{otherwise}.
\end{cases}
\end{equation}
Numeric equivalence is determined by parsing a single numeric value from both the extracted answer and the target; partial credit is awarded only when both parses succeed and the resulting numbers match.

\paragraph{Math-500: accuracy reward.}
For Math-500, we compute correctness by parsing both the model completion and the reference solution into normalized math expressions and verifying semantic equivalence. We parse the gold solution using a math-expression extractor (first-match mode). When the gold solution is parseable, we parse the model completion using a stricter math-expression normalization configuration that rejects malformed operators and enables basic normalization (e.g., equations, units), with boxed extraction prioritized. We then apply a verifier to compare the parsed prediction against the parsed gold expression.
The accuracy reward is
\begin{equation}
r_{\text{acc}} =
\begin{cases}
2.0, & \text{if } \texttt{verify}(\hat{y}, y)=\text{true},\\
0.0, & \text{otherwise}.
\end{cases}
\end{equation}
where $\hat{y}$ and $y$ denote the parsed prediction and gold expressions, respectively. If either parsing or verification fails, we assign reward $0$ for that sample. If the gold solution itself is not parseable, we assign a constant reward of $1.0$ and skip the example to avoid injecting label noise from unparseable references. This accuracy reward follows prior work \cite{lin2025cppo}.

\paragraph{Math-500: format reward.}
We additionally use a binary format reward that encourages presenting the final answer in a boxed form.
The format reward checks whether the completion contains a substring matching \texttt{\textbackslash boxed\{...\}} and is defined as
\begin{equation}
r_{\text{fmt}} =
\begin{cases}
1.0, & \text{if the completion contains } \texttt{\textbackslash boxed\{...\}},\\
0.0, & \text{otherwise}.
\end{cases}
\end{equation}
When using a combined reward, we optimize $r = r_{\text{acc}} + r_{\text{fmt}}$.

\section{Detailed explanation on removing the median completion}
\label{sec:appendix_remove_median}

We show that excluding the median completion does not change the policy-gradient. Our derivation follows the PPO-style clipped surrogate used in CPPO (Eq.~(5)--(6) in \cite{lin2025cppo}), adapted to our $G{+}1$ rollouts and median-centered advantages.

\paragraph{Clipped surrogate and its gradient.}
Omitting the KL term for clarity, the reward-driven surrogate objective is
\begin{equation}
\small
\mathcal{J}(\theta)
=
\mathbb{E}\Bigg[
\frac{1}{G}\sum_{i=1}^{G+1}\frac{1}{|o_i|}\sum_{t=1}^{|o_i|}
\min\Big(
\rho_{i,t}(\theta)\,A_i,\;
\hat\rho_{i,t}(\theta)\,A_i
\Big)
\Bigg],
\label{eq:mcgrpo_obj_Gp1}
\end{equation}
where
$\rho_{i,t}(\theta)=\pi_{\theta}(o_{i,t}\mid q,o_{i,<t})/\pi_{\theta_{\mathrm{old}}}(o_{i,t}\mid q,o_{i,<t})$
and
$\hat\rho_{i,t}(\theta)=\mathrm{clip}(\rho_{i,t}(\theta),1-\epsilon,1+\epsilon)$.
Using $\nabla_\theta \rho_{i,t}(\theta)=\rho_{i,t}(\theta)\nabla_\theta\log\pi_\theta(o_{i,t}\mid q,o_{i,<t})$, the corresponding policy-gradient form (same structure as \cite{lin2025cppo}, Eq.~(6)) can be written compactly as
\begin{equation}
\small
\nabla_{\theta}\mathcal{J}(\theta)
=
\mathbb{E}\Bigg[
\frac{1}{G}\sum_{i=1}^{G+1}\frac{1}{|o_i|}\sum_{t=1}^{|o_i|}
\nabla_\theta \min\!\Big(
\rho_{i,t}(\theta)\,A_i,\;
\hat\rho_{i,t}(\theta)\,A_i
\Big)
\Bigg].
\label{eq:mcgrpo_grad_minform}
\end{equation}

\paragraph{Dropping the median completion.}
In MC-GRPO, we sample one additional rollout, yielding $G{+}1$ completions, and let $i^\star$ satisfy $r_{i^\star}=b(q)$. By Eq.~\eqref{eq:mcgrpo_adv}, $A_{i^\star}=0$.
Define $\mathcal{I}(q)=\{1,\dots,G+1\}\setminus\{i^\star\}$ (Eq.~\eqref{eq:mcgrpo_indexset}). Splitting the sum in Eq.~\eqref{eq:mcgrpo_grad_minform} gives
\begin{equation}
\small
\nabla_{\theta}\mathcal{J}(\theta)
=
\mathbb{E}\Bigg[
\frac{1}{G}\Bigg(
\sum_{i\in\mathcal{I}(q)}\frac{1}{|o_i|}\sum_{t=1}^{|o_i|}
\nabla_\theta \min\!\Big(
\rho_{i,t}(\theta)\,A_i,\;
\hat\rho_{i,t}(\theta)\,A_i
\Big)
\;+\;
\frac{1}{|o_{i^\star}|}\sum_{t=1}^{|o_{i^\star}|}
\nabla_\theta \min\!\Big(
\rho_{i^\star,t}(\theta)\,A_{i^\star},\;
\hat\rho_{i^\star,t}(\theta)\,A_{i^\star}
\Big)
\Bigg)
\Bigg].
\label{eq:mcgrpo_grad_split_clip}
\end{equation}
Since $A_{i^\star}=0$, we have $\rho_{i^\star,t}(\theta)A_{i^\star}=0$  for all $t$, hence the second term vanishes. Therefore, excluding the median completion leaves the gradient unchanged while keeping the effective update size fixed at $G$ trajectories per prompt.
Therefore,
\begin{equation}
\small
\nabla_{\theta}\mathcal{J}(\theta)
=
\mathbb{E}\Bigg[
\frac{1}{G}
\sum_{i\in\mathcal{I}(q)}\frac{1}{|o_i|}\sum_{t=1}^{|o_i|}
\nabla_\theta \min\!\Big(
\rho_{i,t}(\theta)\,A_i,\;
\hat\rho_{i,t}(\theta)\,A_i
\Big)
\Bigg],
\label{eq:mcgrpo_grad_nomedian}
\end{equation}
which is exactly the gradient obtained by excluding the median completion while keeping $|\mathcal{I}(q)|=G$.

\section{Performance in Table Format}
\label{sec:detailed_number}

Tables~\ref{tab:dapo_results}, and \ref{tab:drgrpo_results} report detailed accuracy results under different rollout budgets $G$ across five model--dataset settings on GSM8K and Math-500. Overall, median-centered variants consistently improve upon their corresponding baselines under the same training-sample budget per prompt, with the largest gains typically appearing in the small-rollout regime.
Across GRPO, DAPO, and DR-GRPO, the benefits of median-centering become more pronounced as $G$ decreases, reflecting increased noise in group statistics when only a few rollouts are available. While improvements are generally consistent, the magnitude varies by model and dataset, with occasional near-zero or slightly negative changes in isolated configurations, indicating that effect size depends on the baseline variance and reward landscape.

\begin{table*}[t]
  \centering
\caption{{Results under different DAPO rollout budgets.}
We report final accuracy as a function of the rollout budget $G$ across five model--dataset settings (GSM8K and Math-500).
MC-DAPO constructs a median baseline using additional sampled completions and removes the median (zero-advantage) completion so that the number of completions contributing to the policy-gradient remains $G$, matching DAPO.
Green numbers indicate absolute accuracy improvements over DAPO at the same number of training samples per prompt.}
\label{tab:dapo_results}
  \vspace{-1mm}
  \small
  \setlength{\tabcolsep}{4pt}
  \renewcommand{\arraystretch}{1.05}

  \resizebox{0.95\textwidth}{!}{%
  \begin{tabular}{l| c |l c c l}
    \toprule
    Model & Dataset & Method & \#Rollout ($G$) & \#Training samples & Accuracy (\%) \\
     \midrule

    \multirow{6}{*}{Qwen3-1.7B} & \multirow{6}{*}{GSM8K}
      & DAPO & 8 & 8 & 84.45 \\
    & & \mcshade{MC-DAPO (ours)} & \mcshade{8+1} & \mcshade{8} & \mcshade{85.53~\gain{(+1.08)}} \\
    & & DAPO & 4  & 4 & 79.83 \\
    & & \mcshade{MC-DAPO (ours)} & \mcshade{4+1} & \mcshade{4} & \mcshade{86.00~\gain{(+6.17)}} \\
    & & DAPO & 2  & 2 & 78.08 \\
    & & \mcshade{MC-DAPO (ours)} & \mcshade{3+1} & \mcshade{2} & \mcshade{85.97~\gain{(+7.89)}} \\
     \midrule

    \multirow{6}{*}{Llama-3.2-3B-Instruct} & \multirow{6}{*}{GSM8K}
      & DAPO & 8 & 8 & 80.21 \\
    & & \mcshade{MC-DAPO (ours)} & \mcshade{8+1} & \mcshade{8} & \mcshade{80.23~\gain{(+0.02)}} \\
    & & DAPO & 4  & 4 & 78.69 \\
    & & \mcshade{MC-DAPO (ours)} & \mcshade{4+1} & \mcshade{4} & \mcshade{80.43~\gain{(+1.74)}} \\
    & & DAPO & 2  & 2 & 75.66 \\
    & & \mcshade{MC-DAPO (ours)} & \mcshade{3+1} & \mcshade{2} & \mcshade{77.63~\gain{(+1.97)}} \\
     \midrule

    \multirow{6}{*}{Qwen2.5-Math-1.5B} & \multirow{6}{*}{Math-500}
      & DAPO & 8 & 8 & 70.20 \\
    & & \mcshade{MC-DAPO (ours)} & \mcshade{8+1} & \mcshade{8} & \mcshade{70.60~\gain{(+0.40)}} \\
    & & DAPO & 4  & 4 & 69.40 \\
    & & \mcshade{MC-DAPO (ours)} & \mcshade{4+1} & \mcshade{4} & \mcshade{70.20~\gain{(+0.80)}} \\
    & & DAPO & 2  & 2 & 64.20 \\
    & & \mcshade{MC-DAPO (ours)} & \mcshade{2+1} & \mcshade{2} & \mcshade{65.80~\gain{(+1.60)}} \\
     \midrule

    \multirow{6}{*}{Qwen3-4B-Instruct} & \multirow{6}{*}{Math-500}
      & DAPO & 8 & 8 & 81.80 \\
    & & \mcshade{MC-DAPO (ours)} & \mcshade{8+1} & \mcshade{8} & \mcshade{83.20~\gain{(+1.40)}} \\
    & & DAPO & 4  & 4 & 79.60 \\
    & & \mcshade{MC-DAPO (ours)} & \mcshade{4+1} & \mcshade{4} & \mcshade{82.90~\gain{(+3.30)}} \\
    & & DAPO & 2  & 2 & 77.60 \\
    & & \mcshade{MC-DAPO (ours)} & \mcshade{2+1} & \mcshade{2} & \mcshade{83.00~\gain{(+5.40)}} \\
     \midrule

    \multirow{6}{*}{Qwen2.5-7B-Instruct} & \multirow{6}{*}{Math-500}
      & DAPO & 8 & 8 & 75.80 \\
    & & \mcshade{MC-DAPO (ours)} & \mcshade{8+1} & \mcshade{8} & \mcshade{76.50~\gain{(+0.70)}} \\
    & & DAPO & 4  & 4 & 75.80 \\
    & & \mcshade{MC-DAPO (ours)} & \mcshade{4+1} & \mcshade{4} & \mcshade{77.00~\gain{(+1.20)}} \\
    & & DAPO & 2  & 2 & 74.60 \\
    & & \mcshade{MC-DAPO (ours)} & \mcshade{2+1} & \mcshade{2} & \mcshade{76.00~\gain{(+1.40)}} \\

    \bottomrule
  \end{tabular}%
  }
  \vspace{-2mm}
\end{table*}

\begin{table*}[t]
  \centering
\caption{{Results under different DR-GRPO rollout budgets.}
We report final accuracy as a function of the rollout budget $G$ across five model--dataset settings (GSM8K and Math-500).
MC-DR-GRPO constructs a median baseline using additional sampled completions and removes the median (zero-advantage) completion so that the number of completions contributing to the policy-gradient remains $G$, matching DR-GRPO.
Green numbers indicate absolute accuracy improvements over DR-GRPO at the same number of training samples per prompt.}
\label{tab:drgrpo_results}
  \vspace{-1mm}
  \small
  \setlength{\tabcolsep}{4pt}
  \renewcommand{\arraystretch}{1.05}

  \resizebox{0.95\textwidth}{!}{%
  \begin{tabular}{l| c |l c c l}
    \toprule
    Model & Dataset & Method & \#Rollout ($G$) & \#Training samples & Accuracy (\%) \\
     \midrule

    \multirow{6}{*}{Qwen3-1.7B} & \multirow{6}{*}{GSM8K}
      & DR-GRPO & 8 & 8 & 84.68 \\
    & & \mcshade{MC-DR-GRPO (ours)} & \mcshade{8+1} & \mcshade{8} & \mcshade{87.33~\gain{(+2.65)}} \\
    & & DR-GRPO & 4  & 4 & 83.24 \\
    & & \mcshade{MC-DR-GRPO (ours)} & \mcshade{4+1} & \mcshade{4} & \mcshade{86.35~\gain{(+3.11)}} \\
    & & DR-GRPO & 2  & 2 & 79.15 \\
    & & \mcshade{MC-DR-GRPO (ours)} & \mcshade{3+1} & \mcshade{2} & \mcshade{85.36~\gain{(+6.21)}} \\
     \midrule

    \multirow{6}{*}{Llama-3.2-3B-Instruct} & \multirow{6}{*}{GSM8K}
      & DR-GRPO & 8 & 8 & 79.37 \\
    & & \mcshade{MC-DR-GRPO (ours)} & \mcshade{8+1} & \mcshade{8} & \mcshade{79.98~\gain{(+0.61)}} \\
    & & DR-GRPO & 4  & 4 & 78.77 \\
    & & \mcshade{MC-DR-GRPO (ours)} & \mcshade{4+1} & \mcshade{4} & \mcshade{80.51~\gain{(+1.74)}} \\
    & & DR-GRPO & 2  & 2 & 75.74 \\
    & & \mcshade{MC-DR-GRPO (ours)} & \mcshade{3+1} & \mcshade{2} & \mcshade{79.50~\gain{(+3.76)}} \\
     \midrule

    \multirow{6}{*}{Qwen2.5-Math-1.5B} & \multirow{6}{*}{Math-500}
      & DR-GRPO & 8 & 8 & 69.60 \\
    & & \mcshade{MC-DR-GRPO (ours)} & \mcshade{8+1} & \mcshade{8} & \mcshade{71.40~\gain{(+1.80)}} \\
    & & DR-GRPO & 4  & 4 & 67.60 \\
    & & \mcshade{MC-DR-GRPO (ours)} & \mcshade{4+1} & \mcshade{4} & \mcshade{70.80~\gain{(+3.20)}} \\
    & & DR-GRPO & 2  & 2 & 63.80 \\
    & & \mcshade{MC-DR-GRPO (ours)} & \mcshade{2+1} & \mcshade{2} & \mcshade{67.60~\gain{(+3.80)}} \\
     \midrule

    \multirow{6}{*}{Qwen3-4B-Instruct} & \multirow{6}{*}{Math-500}
      & DR-GRPO & 8 & 8 & 81.40 \\
    & & \mcshade{MC-DR-GRPO (ours)} & \mcshade{8+1} & \mcshade{8} & \mcshade{83.40~\gain{(+2.00)}} \\
    & & DR-GRPO & 4  & 4 & 79.80 \\
    & & \mcshade{MC-DR-GRPO (ours)} & \mcshade{4+1} & \mcshade{4} & \mcshade{84.10~\gain{(+4.30)}} \\
    & & DR-GRPO & 2  & 2 & 78.60 \\
    & & \mcshade{MC-DR-GRPO (ours)} & \mcshade{2+1} & \mcshade{2} & \mcshade{83.80~\gain{(+5.20)}} \\
     \midrule

    \multirow{6}{*}{Qwen2.5-7B-Instruct} & \multirow{6}{*}{Math-500}
      & DR-GRPO & 8 & 8 & 76.20 \\
    & & \mcshade{MC-DR-GRPO (ours)} & \mcshade{8+1} & \mcshade{8} & \mcshade{77.60~\gain{(+1.40)}} \\
    & & DR-GRPO & 4  & 4 & 76.00 \\
    & & \mcshade{MC-DR-GRPO (ours)} & \mcshade{4+1} & \mcshade{4} & \mcshade{75.90~\gain{(-0.10)}} \\
    & & DR-GRPO & 2  & 2 & 73.00 \\
    & & \mcshade{MC-DR-GRPO (ours)} & \mcshade{2+1} & \mcshade{2} & \mcshade{76.20~\gain{(+3.20)}} \\

    \bottomrule
  \end{tabular}%
  }
  \vspace{-2mm}
\end{table*}


\end{document}